# A Dashboard to Analysis and Synthesis of Dimensionality Reduction Methods in Remote Sensing

Elkebir Sarhrouni [#1], Ahmed Hammouch [*2], Driss Aboutajdine [#3]

[#]LRIT, Faculty of Sciences, Mohamed V - Agdal University, Morocco,
A. Ibn Battouta. 4. B.P. : 1014 Rabat, Morocco
[1] sarhrouni436@yahoo.fr
[*] LRGE, ENSET, Mohamed V - Souissi University, Morocco
A. FAR BP : 6207 - Rabat Instituts Rabat Morocco

*Abstract*—Hyperspectral images (HSI) classification is a high technical remote sensing software. The purpose is to reproduce a thematic map . The HSI contains more than a hundred hyperspectral measures, as bands (or simply images), of the concerned region. They are taken at neighbors frequencies. Unfortunately, some bands are redundant features, others are noisily measured, and the high dimensionality of features made classification accuracy poor. The problematic is how to find the good bands to classify the regions items. Some methods use Mutual Information (MI) and thresholding, to select relevant images, without processing redundancy. Others control and avoid redundancy. But they process the dimensionality reduction, some times as selection, other times as wrapper methods without any relationship . Here , we introduce a survey on all scheme used, and after critics and improvement, we synthesize a dashboard, that helps user to analyze an hypothesize features selection and extraction softwares.

Keyword-Feature Selection Software, Feature Extraction Software, Hyperspectral images Classification, Remote Sensing.

I. INTRODUCTION

Due to the recent achievements in the remote sensing technologies, we are faced at large quantity of information, organized at bidirectional measures of the same region, called bands, and taken at very closed frequencies. The goal here is to determine patterns in order to classify the points and produce the thematic map of the concerned region. This technology is called Hyperspectral Images (HSI), and it opens new applications fields and renews the problematics posed in classification domain. Explicitly we are faced at reduction of dimensionality problematic. the feature classification domain, the choice of data affects widely the results.So, the bands don't all contain the information; some ands are irrelevant like those affected by various atmospheric effects,and decrease the classification accuracy. And there exist redundant ands to complicate the learning system and product incorrect prediction [1].

Even the bands contain enough information about the scene they may can't predict the classes correctly if the dimension of space images, is so large that needs many cases to detect the relationship between the bands and the scene (Hughes phenomenon) [5]. We can reduce the dimensionality of hyperspectral images by selecting only the relevant bands (feature selection or subset selection methodology), or extracting, from the original bands, new bands containing the maximal information about the classes, using any functions, logical or numerical (feature extraction methodology) [4,6], or we can use an hybrid schema containing selection before extraction.

An example of Hyperspectral image that largely served for academic search is AVIRIS 92AV3C (Airborne Visible Infrared Imaging Spectrometer). [2]. It contains 220 images taken of the region "Indiana Pine" at "north-western Indiana", USA [2]. The 220 called bands are taken between 0.4µm and 2.5µm. Each band has 145 lines and 145 columns. The ground truth map is also provided, but only 10366 pixels are labeled fro 1 to 16. Each label indicates one from 16 classes. The zeros indicate pixels how are not classified yet, see Figure.1.

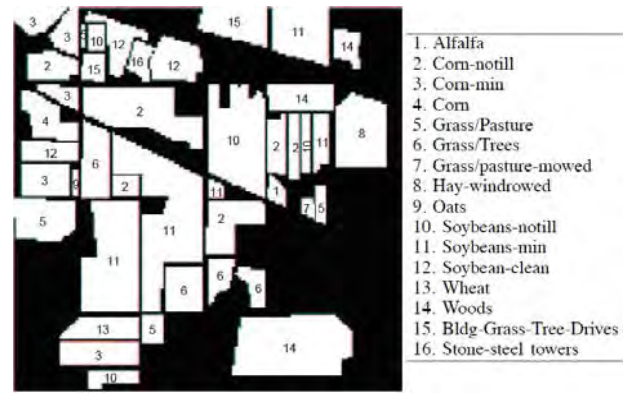

Fig. 1. The Ground Truth map of AVIRIS 92AV3C and the 16 classes.

The hyperspectral image AVIRIS 92AV3C contains numbers (measures) between 955 and 9406. Each pixel of the ground truth map has a set of 220 numbers along the hyperspectral image. This numbers (measures) represent the reflectance of the pixel in each band.

So reducing dimensionality means selecting only the dimensions caring a lot of information regarding the classes.

In this paper we collect all techniques and algorithms related to dimensionality reduction applied to Hyperspectral images, or that can be applied to HSI. This survey provides a work board for analysis and synthesis of dimensionality reduction.

We start, in section two by citing the principle notions related at classification in the context of dimensionality reduction. In section three, we develop different schema and strategies for features selection and extraction, and we discuss their ability to be applied to HSI. The forth section provides our synthesis of a work board flow for selection, extraction and classification of HSI remote sensing

## II. DIMENSIONALITY REDUCTION

*A.    General Purpose*

AN important question that often arises in the field of data mining is the problem of having too many attributes. In other words, the measured attributes are not all likely to be necessary for accurate discrimination. Therefore include some features the model classification can lead to a worse model than if they were removed. In addition it is not clear if the features are (or are not) relevant. In this context, dimensionality reduction use data-adaptive methods using a priori information available, to inform us which variables are clearly relevant or not to the task of classification [3].

*B.    The Curse of Dimensionality*

While, theoretically, having more features should give us more power to discriminate classes, the real world gives us many reasons why this is not usually the case. This means that what may work well in a unidimensional may not be extended to high dimensions. This is the case where the amount of data must be increased to maintain a level of precision parameters.

Thus, in a task induction, when the number of attributes increases, the time required for an algorithm sometimes grows exponentially. Therefore, when the set of attributes is large enough, the induction algorithms are simply insurmountable. So that the induction methods suffer from the curse of dimensionality [5].

This problem is compounded by the fact that many attributes can be either irrelevant or redundant with other features in predicting the class of an instance. In this context, these attributes serve no purpose except to increase the s induction time.

*C.    Generic Process for Dimensionality Reduction*

The dimensionality reduction techniques usually involve of both search algorithms and scoring functions. Search algorithms generate subsets solutions possible, and they compare them by using the score as a measure of the effectiveness of each solution. But the search for optimal solution is unattainable due to calculation cost. Aha [9] cites three typical components to achieve a dimensionality reduction of features: a search algorithm to traverse the space of subsets possible, an evaluation function to maximize, receiving a subset of features and provides a numeric value; and finally a performance function which is here the classification.

*D.    Categorization of Dimensionality Reduction Methods According to the Feature Generation Process*

Seen from the generation attributes, dimensionality reduction methods are either:

- ✓    They realize the reduction by transformation of data vectors which gives the attribute extraction,

- ✓ Or without transformation vector data, which is a selection of attributes,
- ✓ Or a selection followed by feature extraction .

E.  *Summary of Dimensionality Reduction Steps*

We reserve the term "selection" to the "generation" subset as the same order as the terms "extraction". And we describe the four stages of the scheme Dash [12] as four steps of dimensionality reduction that can be applied with both the selection and extraction, Figure.2. illustrates this scheme.

More generally, the idea of incorporating four stages Dash [12], we use these steps for the scheme overall dimensionality reduction:

- ✓ A procedure for generating candidate subsets by selection, extraction or extraction followed by selection,
- ✓ An evaluation function to evaluate this subset,
- ✓ A stopping criterion for deciding when to stop the search,
- ✓ A validation procedure to determine the subset keep or abandon it.

F.  *Filter Strategy*

Dimensionality reduction attribute is seen as a step pretreatment. The major drawback of this approach is that it completely ignores the effect of selected attributes on the performance of the induction algorithm.

Figure.3. describes the model selection attribute called Filter (Filter approach) which characterizes these algorithms.

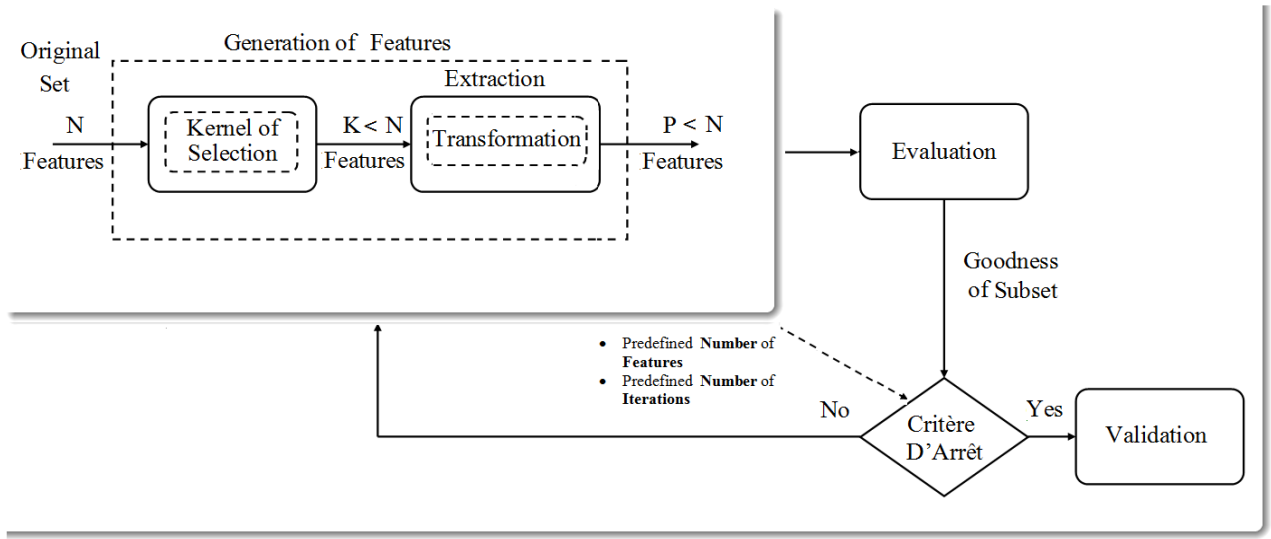

Fig. 2.Dimensionality Reduction in Four Steps

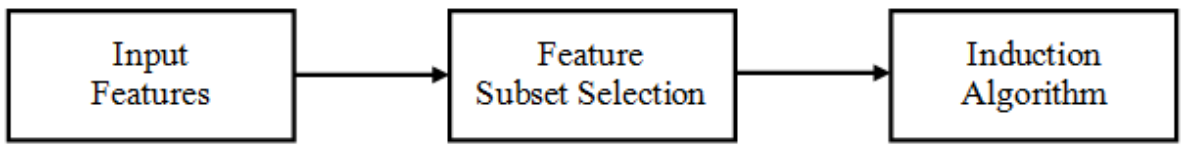

Fig. 3.Dimensionality Reduction : Filter approach

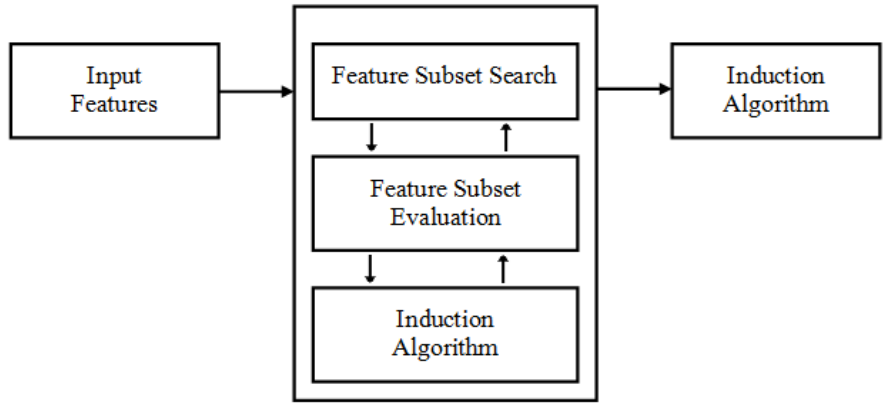

Fig. 4.Wrapper Model : The Induction Algorithm is used as Black Box by Selection Algorithm (From George, 94) )

George [4] states that the selected attributes are also dependent on the induction algorithm. They argue that the Filter approach should be replaced by Wrapper approach.

*G. Wrapper Strategy*

In the model called "Wrapper", the induction algorithm itself is used to determine the relevance of an attribute. The problem is formulated as: Given a subset of attributes, we estimate the precision of the structure that it induces. Therefore, we evaluated the relevance of attributes selected by cross validation. This approach uses the classifier as a "black box" and have an outer loop (or "wrapper") that systematically adds and subtracts the current set of attributes, each subset being evaluated on the basis of performance product classification model. George [4] indicates that process according to the following Figure.54

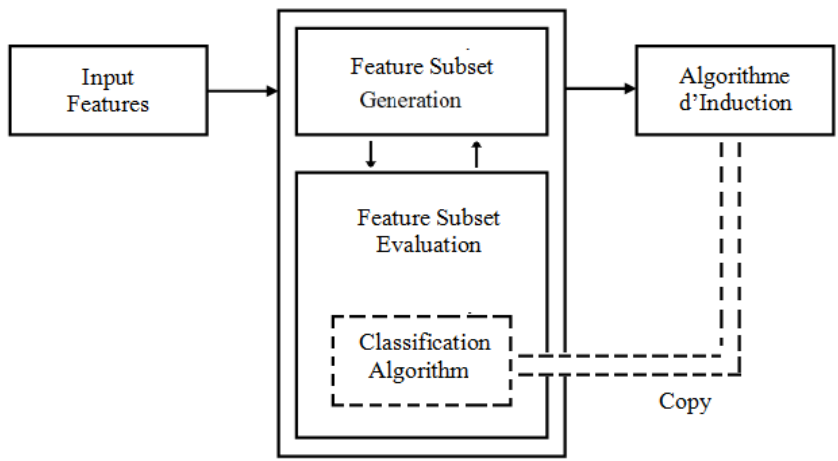

Fig. 5.Modified Scheme of Strategy Wrapper

*H. Hybrid Strategy*

We point out here that Chen [13] offers a combination Hybrid Filter-Wrapper for feature selection. A typical hybrid algorithm, using both an independent measure and a learning algorithm to evaluate subsets of attributes. Independent measurement determines the best subset for a given cardinality. The learning algorithm selects the best subset of the best final subsets found for different cardinalities. The quality of the results of a learning algorithm provides a natural stopping criterion.

*I. Critical of the two Strategy Filter and Wrapper*

We note, according to their scheme,that the wapper strategy differs from the strategy filter by the presence of a loop that uses the "same" induction algorithm (classification) to inform the generation of attributes on the relevance of a subset of candidate attributes. For this, there are models in the literature using an approximation class without using the induction algorithm. Sarhrouni et al. [8] uses an averaging approximation of the truth map for dimensionality reduction and hyperspectral image classification. Demir [7] uses a third strategy called

"Grouping of attributes", he replace a subset of attributes that are highly correlated with their average. We believe that this is a feature extraction. Guo [2] also uses the average (which is a group of attributes) of some bands as aa approximation of the ground truth map.

And in many applications, an approximation of the class (ground truth) can be used in a wrapper strategy, which is not a pure strategy wrapper nor a pure strategy filter, since the algorithm takes into account the effect of selected attributes on the choice of prior validation subset, but we can tell it a " Low Wrapper strategy ".

Note finally that in the scheme of George evaluation subsets is done with the induction algorithm, and hence, it should be noted that the block induction algorithm of evaluation should be included in the evaluation. and the selection block must be a generation block to cover the extraction data methods . We then propose changes presented in Figure.5.

We propose a categorization over more general form "Mixed Strategy Filter-Wrapper" .

*J. Mixed Strategy Filter-Wrapper*

We propose to take into account that the induction algorithm itself is NOT used to decide about the relevance of an attribute. We add that we must consider the degrees of membership of any method of the two strategies on filter and wrapper. For example, if the feedback is achieved by averaging [3,8] the selected attributes, then the algorithm in question is close to the filter model. If the feedback is produced for example by a supervised classifier or not, or another technique (like KNN) is used to approximate the class (ie the truth field), then the method in question is closer to the model Wrapper.

In this vision can be described a model Pure Wrapper using the induction algorithm for feedback. And we can call Pure Filter algorithm using no feedback (measure of relevance of the candidate subset for classification) as shown in the diagram in Figure.6.

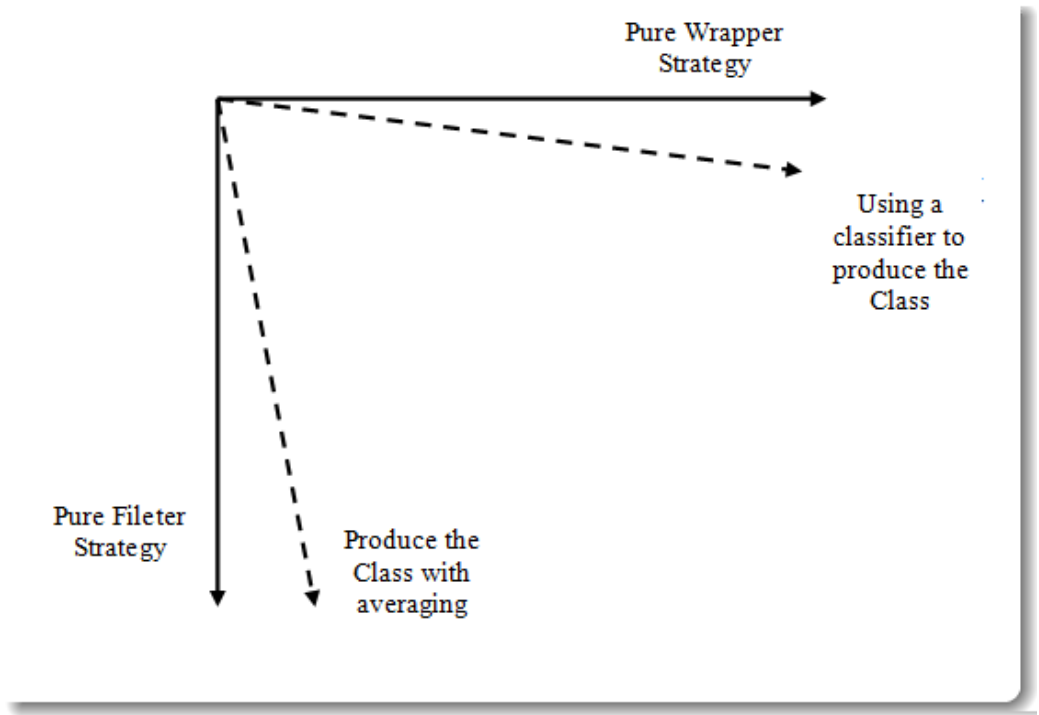

Fig. 6.Principle of Proposed Scheme Mixed Filter-Wrapper

The problem is formulated as follows: given a subset of attributes, we want to estimate the accuracy of their structure induced. Therefore, we evaluated the relevance of attributes selected by its ability to estimate an "estimated" truth map, nearest than the class (the real ground truth) with the same measures

The model of the mixed strategy can represent both the Filter strategy and the Wrapper strategy : Filter Model is a Mixed strategy with feedback using no classifier. Wrapper model is a mixed strategy as feedback is made of the same classifier used for the classification task. We then proposed to represent the interaction of the Mixed model with evaluation measures, as shown in Figure.7.

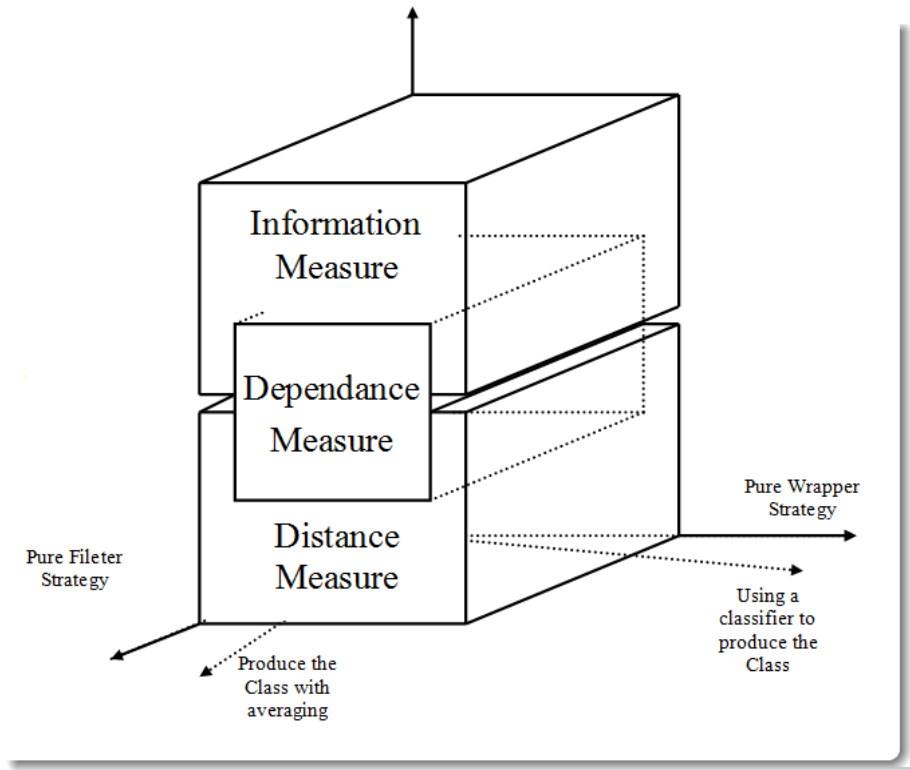

Fig. 7. Interaction of the Proposed Mixed Model with Measures Evaluation

Example of using the diagram to analyze an algorithm: Guo[3,7] uses both a Filter and a weighting kernel SVM classifier by mutual information in the selection and classification of hyperspectral images. The diagram proposed allows us to locate the operating point of the algorithm, and its situates such an application in a broader space.

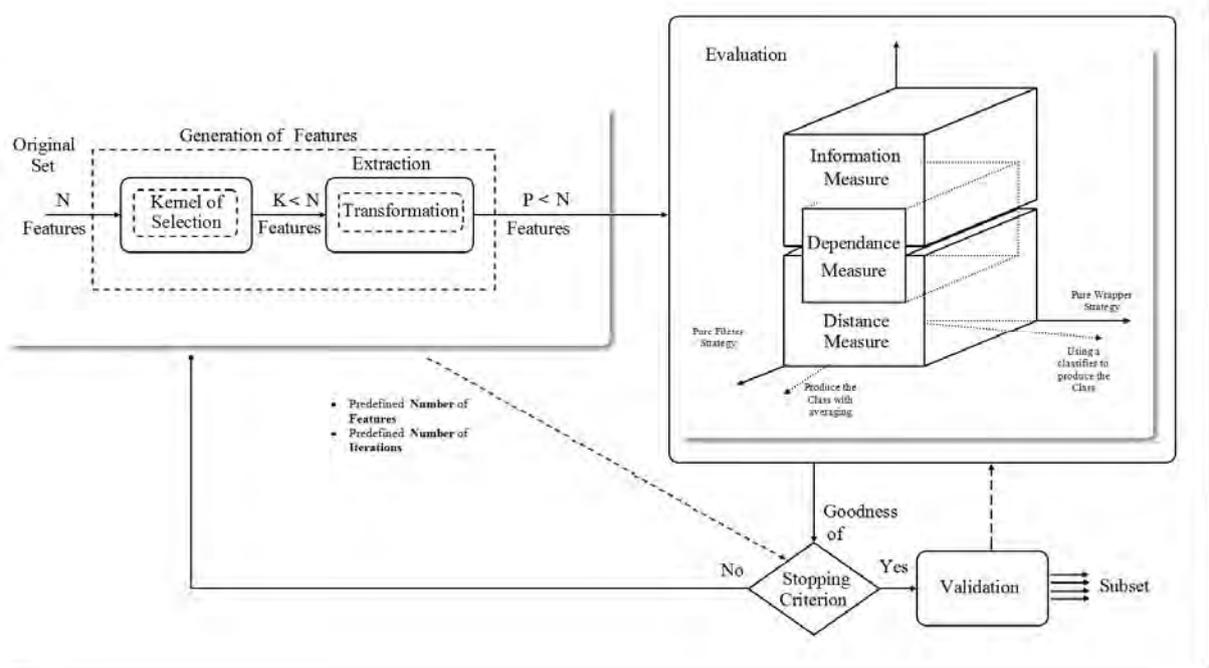

Fig. 8. Proposed Dashboard For Analysis and Synthesis of Dimensionality Reduction Algorithms

III. CONCLUSION

In this paper, we had treat the strategy of dimensionality reduction, and specially for remote sensing technics, as hyperspectral images. we find that the strategies in the literature aren't able to make analyze correctly the existent features selection and extraction algorithms. So we had improved some scheme, and let them able to be more general. We had extend some notions to be more expressive, for interpret the dimensionality reduction algorithms. and finally we synthesize a new scheme as dashboard for analyze and synthesize dimensionality reduction software.